\documentclass{article}

\usepackage[main, final]{neurips_2026}

\usepackage[utf8]{inputenc} 
\usepackage[T1]{fontenc}    
\usepackage{hyperref}       
\usepackage{url}            
\usepackage{booktabs}       
\usepackage{amsfonts}       
\usepackage{nicefrac}       
\usepackage{microtype}      
\usepackage{xcolor}         
\usepackage{amsmath} 
\usepackage{graphicx}
\usepackage{subcaption}
\usepackage{wrapfig}
\usepackage{pifont}
\usepackage{colortbl}
\usepackage{multirow}
\usepackage{xcolor}
\usepackage[skins,breakable]{tcolorbox}
\usepackage{enumitem}
\usepackage[dvipsnames]{xcolor}
\usepackage{hyperref}
\newcommand{\pf}[1]{\cellcolor{red!30}#1}
\newcommand{\ps}[1]{\cellcolor{orange!30}#1}
\newcommand{\pt}[1]{\cellcolor{yellow!30}#1}

\title{Morph$\mathcal{S}\mathcal{I}\mathcal{G}$: $\mathcal{S}$ubject-Driven $\mathcal{I}$mage $\mathcal{G}$eneration via Decoupled Anchoring and Feature Transport}

\author{%
  Hailong Yan$^1$\thanks{Project Leader.} \quad
  Yongrui Zhang$^1$ \quad
  Xiangtao Zhang$^1$ \quad
  Le Zhang$^{2,1}$\thanks{Corresponding author.} \\[1.0ex]
  $^1$ UESTC \qquad\qquad $^2$ Hohai University \\
  \texttt{yanhailong@std.uestc.edu.cn, lezhang@uestc.edu.cn}
}

\begin{document}
\maketitle

\begin{figure*}[htbp]
\centering
\includegraphics[width=\linewidth]{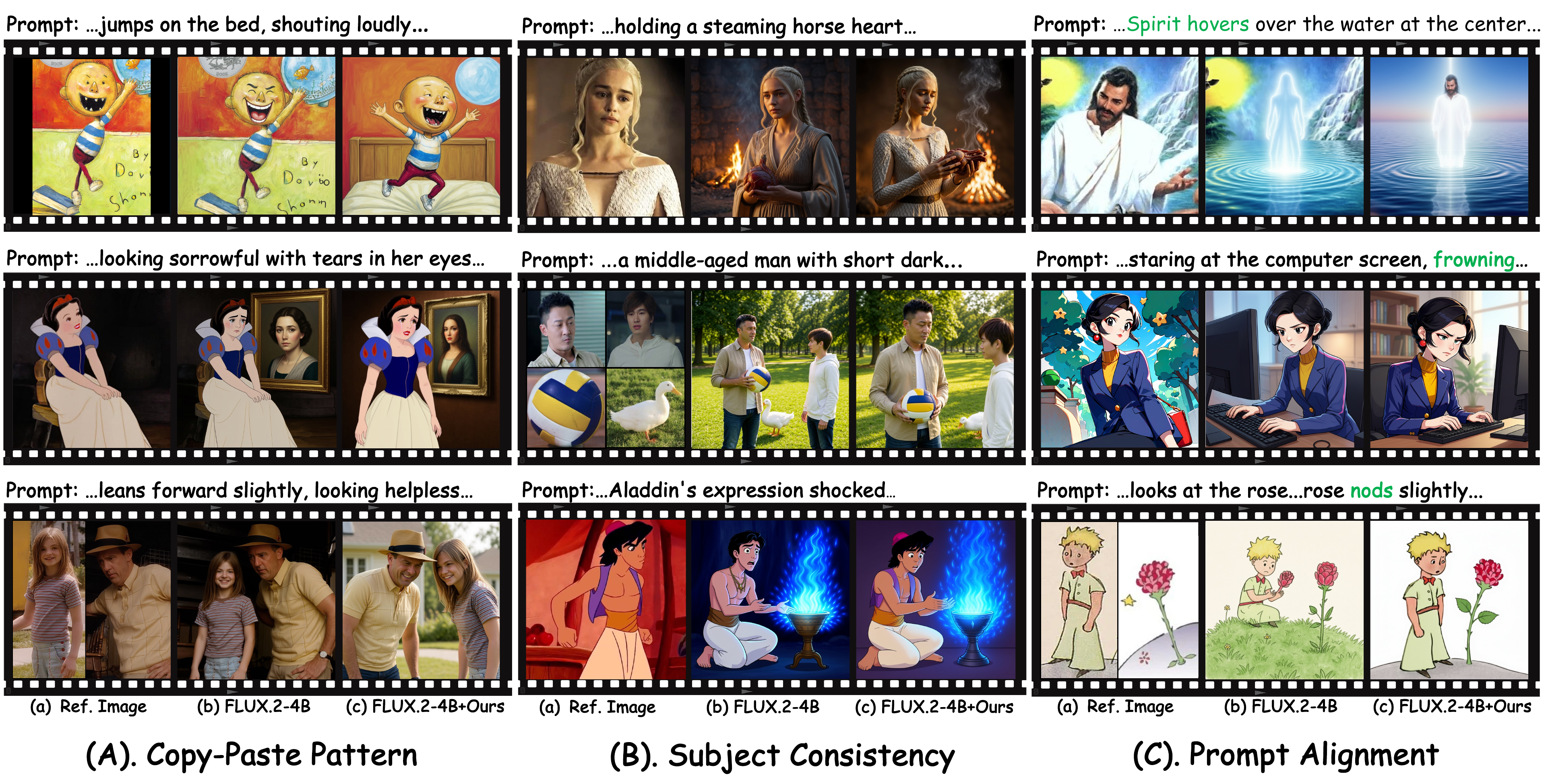}
\caption{\label{fig1} Qualitative comparison between MorphSIG and the baseline.}
\end{figure*}

\begin{abstract}
We propose MorphSIG, a training-free framework that reformulates subject-driven image generation (SIG) as pseudo-video feature transport. MorphSIG contains two core modules: Progressive Decoupled Anchoring, which uses frequency-domain decoupling and structural scrambling during early denoising to break structural locking and enable a gradual transition from free composition to identity alignment; and Pseudo-flow Guided Feature Transport, which constructs a latent-space pseudo-flow field to mimic Image-to-Video temporal coherence and transport subject semantics precisely. To address stylistic homogeneity and the lack of ground truth in existing benchmarks, we further introduce a benchmark with diverse stylized subjects. Extensive experiments show that MorphSIG significantly improves pose diversity and subject fidelity over baselines without additional training. Overall, MorphSIG bridges static generation and dynamic propagation, providing a cost-effective paradigm for high-fidelity subject consistency. The Code and benchmark will be available \href{https://github.com/LukasYan30/MorphSIG}{here}.
\end{abstract}

\section{Introduction}

Subject-driven image generation (SIG) aims to synthesize images that follow text prompts while preserving the details of reference subjects \cite{kumari2023multi,ruiz2023dreambooth,ding2024freecustom,wu2025less,cheng2025umo}. By enabling user-defined customization, SIG has broad applications in film production \cite{yan2026animeagent} and advertising \cite{gao2025postermaker}.

In practical applications, SIG faces two major challenges: \ding{182} the "copy-paste pattern," where generated results inherit the reference pose and lack compositional freedom; and \ding{183} "subject inconsistency," where subject features degrade during diffusion, reducing identity stability.

\begin{wrapfigure}[15]{r}{0.6\textwidth} 
\centering
\vspace{-14pt} 
\includegraphics[width=\linewidth]{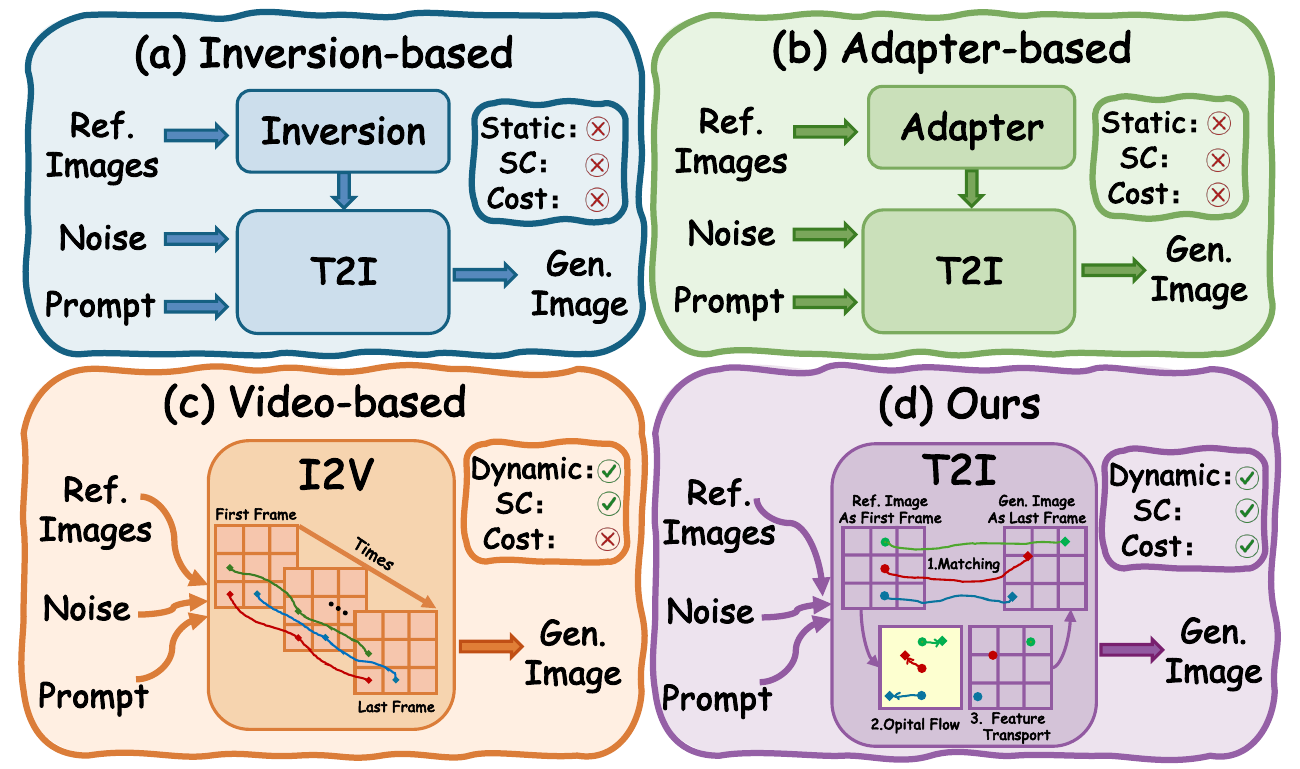}
\vspace{-12pt} 
\centering
\caption{\label{fig2}Comparison of different SIG paradigms. SC (Subject Consistency), and Cost (computational complexity).}
\end{wrapfigure}

As illustrated in Fig.~\ref{fig2}, existing training-based SIG methods anchor identity through large-scale paired datasets \cite{ruiz2023dreambooth, deng2025magref, xu2025withanyone, wu2025less, cheng2025umo} or the introduction of auxiliary encoding networks \cite{ye2023ip, mou2024t2i, wang2024ms}, yet these incur significant computational costs and are prone to overfitting. Training-free approaches \cite{zhang2025freecus, mazuz2025consistyle, feng2026personalize, chan2024improving} typically inject subject features via attention-sharing mechanisms after inverting the reference image; however, they struggle to preserve identity due to the lack of consistency constraints. Another research \cite{yan2026animeagent, wu2025chronoedit, lin2025realgeneral, zhang2025image, tong2026cof, chen2025unireal} attempts to leverage the dynamic priors of video models for image-level SIG, but suffers from low efficiency due to the generation of redundant transition frames.

To address these challenges, we analyze the underlying feature representations and inference paradigms. The "copy-paste pattern" arises from a pronounced "high-frequency shortcut effect" in Text\&Image-to-Image (T\&I2I) models, where high-frequency components from the reference image dominate early denoising and prematurely lock pose evolution. Meanwhile, the superior consistency of video models comes from temporally coherent feature transport, characterized by displacement mapping and semantic propagation across adjacent frames.

To balance fidelity and diversity, we propose MorphSIG, a training-free SIG framework. First, to alleviate the "copy-paste" pattern, we introduce \textbf{Progressive Decoupled Anchoring}. By applying low-pass filtering and high-frequency structural permutation to reference Keys in joint self-attention, it removes global geometric bias while preserving essential feature statistics. Combined with time-dependent anchoring strength, this strategy unlocks compositional flexibility during early diffusion. Second, to improve consistency, we reformulate T\&I2I as a "two-frame video generation" paradigm, treating the reference and target images as pseudo-first and pseudo-last frames. Based on this view, we propose \textbf{Pseudo-flow Guided Warped Feature Transport}, which simulates temporal coherence by warping reference features with pseudo-flow. Combining low-frequency Spherical Linear Interpolation and high-frequency Winner-Take-All enables effective transport of warped reference features, improving detail generation while preserving structural consistency.

We further introduce a benchmark with diverse styles and high-quality Ground Truth, enabling rigorous evaluation of fidelity and diversity in SIG. Our contributions are summarized as follows.

\begin{itemize}

\item We identify the "high-frequency shortcut" as the main cause of the "copy-paste" pattern in T\&I2I, and redefine T\&I2I as a "pseudo-video" task to simulate I2V temporal coherence without redundant computation.

\item We propose Progressive Decoupled Anchoring to mitigate high-frequency shortcuts, and Pseudo-flow Guided Feature Transport to preserve subject consistency.

\item We introduce a new SIG benchmark with diverse styles and high-quality Ground Truth, addressing the lack of objective metrics and stylistic diversity in existing datasets.

\item As a training-free framework, MorphSIG achieves superior fidelity-diversity balance on both public and our proposed benchmarks.

\end{itemize}

\section{Related Work}

\textbf{Consistent Text-to-Image Generation}. While fine-tune approaches \cite{ruiz2023dreambooth,kumari2023multi,gu2023mix,wang2026fair,tang2026evolutionary,bai2026towards} are capable of capturing visual details, they suffer from high computational overhead. To balance efficiency and performance, lightweight adapters \cite{mou2024t2i,zhang2025ip,ye2023ip,yan2025mobileie,yan2025igdnet,yan2026revisiting} inject external guidance into frozen backbone; however, these often lack sufficient representation capacity. In contrast, training-free methods \cite{ding2024freecustom,tewel2024training,mazuz2025consistyle,wang2025characonsist,li2025consistent,pu2026drag,pu2026dragging} leverage in-context learning for "plug-and-play" flexibility, but struggle with complex instructions, often failing to maintain textual consistency under large-scale pose variations.

\textbf{Subject-driven Image Generation} synthesizes reference images to precisely preserve subject identities \cite{wu2025less, cheng2025umo, wu2025multicrafter}. For instance, DreamO \cite{mou2025dreamo} and USO \cite{wu2025uso} integrate reference images to generate outputs that mirror both input identity and aesthetics. While frameworks like OmniGen \cite{xiao2025omnigen, wu2025omnigen2} and others \cite{hu2025dynamicid, she2025mosaic, dalva2025lorashop, tan2025ominicontrol} enhance identity decoupling, they often necessitate large-scale retraining, which limits their scalability. Conversely, training-free methods \cite{ zhang2025freecus, hu2026storytailor, chen2026lamic, kang2025flux} utilize optimized attention mechanisms to inject subject features across frames during inference.

\begin{figure*}[tbp]
\centering
\includegraphics[width=\linewidth]{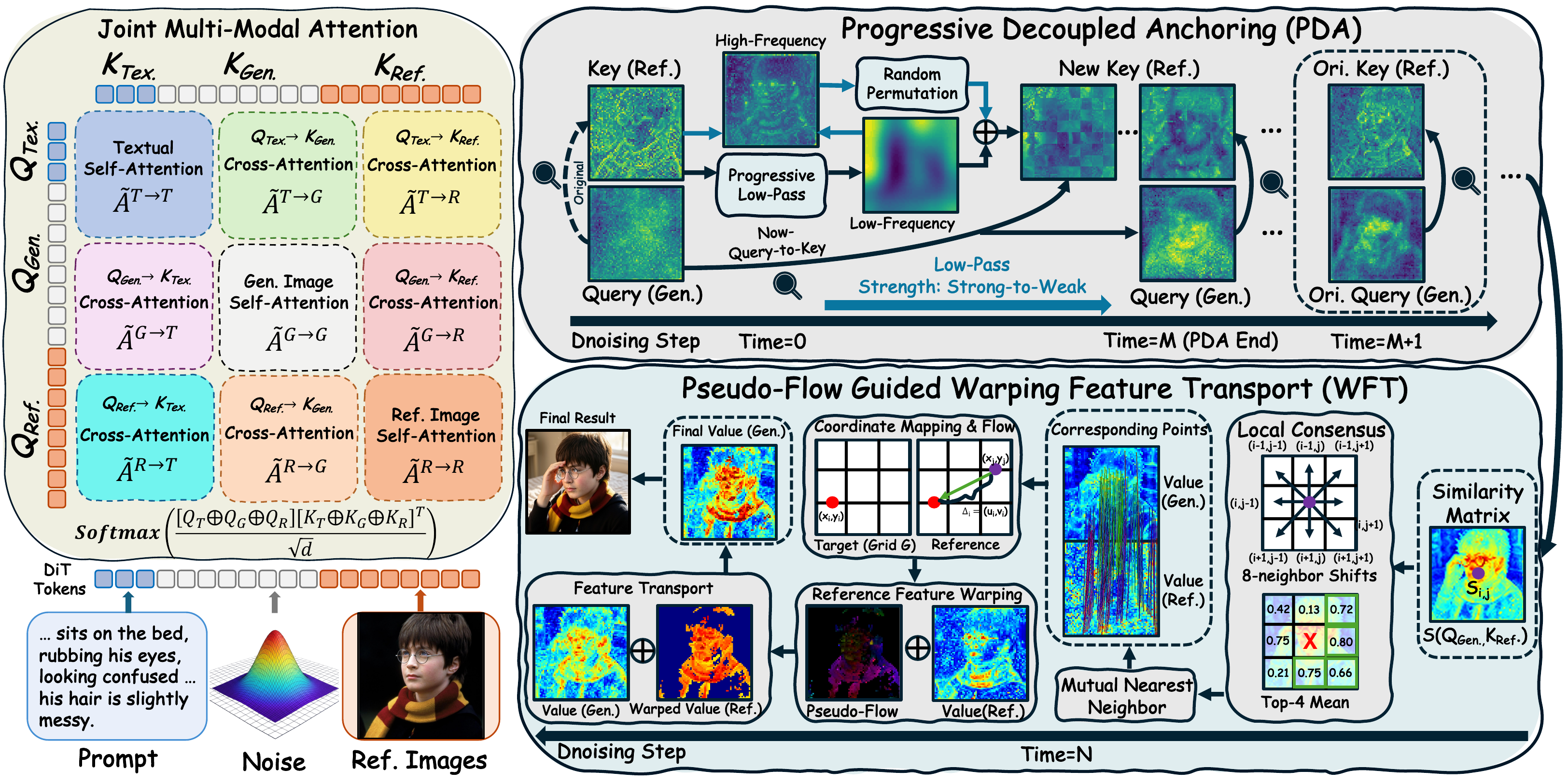}
\caption{\label{figpipeline}Overview of the proposed framework. MorphSIG consists of two modules: Progressive Decoupled Anchoring (PDA) and Warping Feature Transport (WFT). PDA suppresses structural shortcut copying by decoupling and progressively restoring reference anchors during denoising, enabling flexible composition. WFT estimates pseudo-flow to warp and transport reference features into the generation space, improving identity consistency under large pose and geometry variations.}
\vspace{-4pt}
\end{figure*}

\textbf{Image Generation and Video.} Recent video models \cite{blattmann2023stable, blattmann2023align, chen2023videocrafter1, guo2023animatediff} have inspired using video priors for image tasks \cite{rotstein2025pathways, yan2026animeagent, zhang2025image, wu2025chronoedit}. Specifically, UniReal \cite{chen2025unireal} and iMontage \cite{fu2025imontage} treat image generation as video tasks to leverage world dynamics, while RealGeneral \cite{lin2025realgeneral} unifies tasks via conditional frame prediction. To handle complexity, PLACID \cite{tarres2026placid} uses spatio-temporal trajectories for feature constancy, and CoF-T2I \cite{tong2026cof} employs "Chain-of-Frames" for progressive refinement. Despite these gains, video-based models suffer from high overhead and latency. Conversely, we propose achieving high-quality generation by endowing static diffusion models with dynamic evolutionary capabilities.

\section{Method}
\subsection{Preliminaries}
\textbf{Joint Multi-Modal Attention in T2I.} Modern diffusion transformers like FLUX.2 utilizes a Joint Multi-Modal Attention mechanism that unifies text, latent noise, and reference images into a single 1D sequence. The model serializes the text prompt into $X_{\text{txt}}$, the target noise into $X_{\text{gen}}$, and $N$ reference images into a concatenated stream $X_{\text{ref}} = [X_{\text{ref}}^{(1)}; \dots; X_{\text{ref}}^{(N)}]$. To distinguish these "canvases," 4D coordinate IDs are assigned to every token, using discrete temporal offsets ($T$) for each reference image to preserve instance integrity within the unified vector. In the core \textbf{Single-Stream Blocks}, these modalities are fused into a global sequence $X_{\text{all}} = [X_{\text{txt}}; X_{\text{gen}}; X_{\text{ref}}]$, where attention is computed as:
\begin{equation}
\text{Output} = \text{Softmax}\left(\frac{Q K^\top}{\sqrt{d_h}}\right) V, \quad \{Q, K, V\} = \text{Linear}(X_{\text{all}}).
\end{equation}
By treating references as tokens rather than external conditions, the joint attention matrix implicitly facilitates simultaneous self-attention, text-guidance, and multi-source reference injection.

\textbf{3D Attention in I2V.} Video DiT extend static DiTs by processing 4D video tensors $\mathcal{V} \in \mathbb{R}^{f \times c \times h \times w}$ to capture spatiotemporal dynamics. In each layer $l$, video tokens $z_{\text{video}}$ and text tokens $z_{\text{text}}$ are concatenated into an augmented sequence. The queries $\mathbf{Q}_{t,l}$ are formed by merging projections from all $f$ frames and the text prompt: $\mathbf{Q}{t,l} = \text{Concat}(\mathbf{Q}{t,l}^1, \dots, \mathbf{Q}{t,l}^{f}, \mathbf{Q}{t,l}^{\text{text}}).$ The resulting global 3D attention matrix $\mathbf{A}_{t,l} = \text{Softmax}(\mathbf{Q}_{t,l} \mathbf{K}_{t,l}^\top / \sqrt{d})$ facilitates four key interactions: self-frame attention ($\mathbf{A}_{t,l}^{i,i}$) for 2D spatial correlations, cross-frame attention ($\mathbf{A}_{t,l}^{i,j}$) for inter-frame pixel relevance $\text{Attn}(I_i(p) \rightarrow I_j(q))$, and text-frame/self-text attention for semantic guidance. VDTs leverage inherent "world knowledge" to understand complex motion and temporal consistency \cite{nam2025emergent}.

\begin{figure*}[tbp]
\centering
\includegraphics[width=\linewidth]{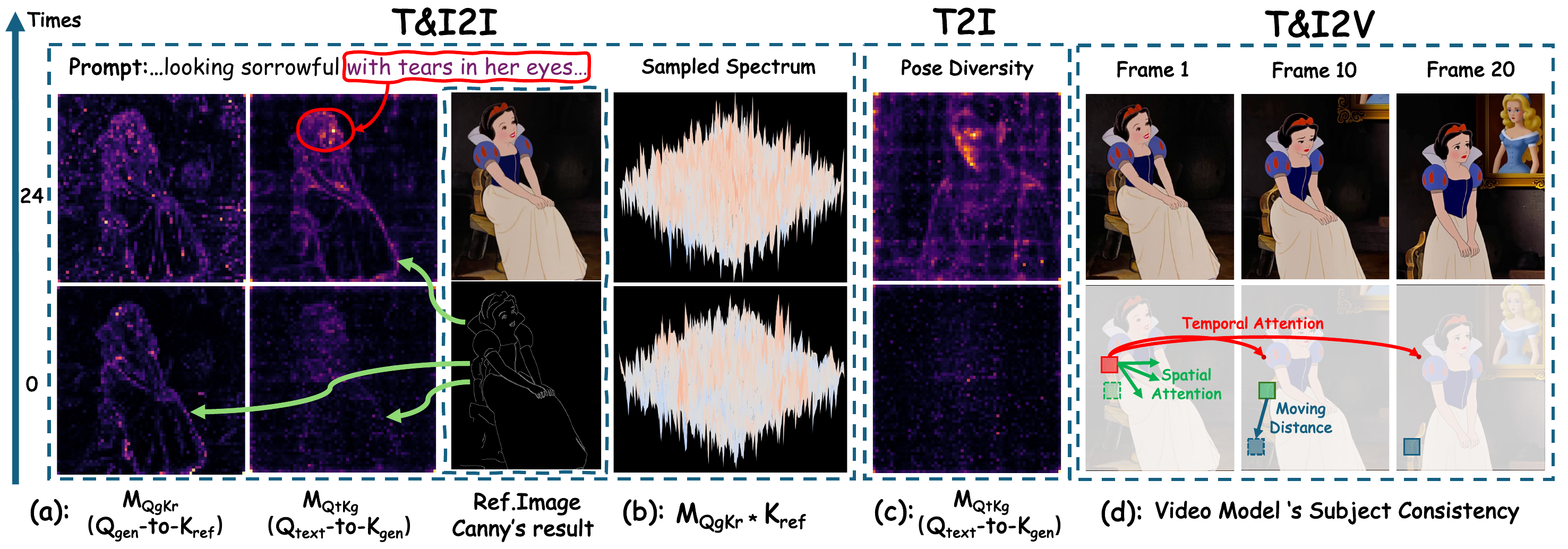}
\caption{\label{fig3} Analysis of High-frequency Shortcut in T\&I2I and Temporal Feature Transport in I2V.}
\vspace{-8pt}
\end{figure*}

\subsection{Problems and Analysis}

We analyze two key observations in T\&I2I models, as shown in Fig.\ref{fig1}: \textbf{Observation \ding{182}: Copy-Paste Pattern.} Generated images rigidly inherit the reference pose. \textbf{Observation \ding{183}: Subject Inconsistency.} Subject identity or features gradually degrade during generation. We first analyze and formally explain these issues, then propose targeted optimizations to balance fidelity and diversity.

\textbf{Analysis for \ding{182}:} \textit{The consistency gain in T\&I2I comes at the expense of generation freedom.}

As shown in Fig.\ref{fig1}(A), T\&I2I models often directly replicate reference features, reducing diversity \cite{goyal2025preventing, li2026disentangling, xu2025withanyone, deng2025magref}. We attribute this to a high-frequency shortcut effect: although diffusion models generally follow a coarse-to-fine generation process \cite{yi2024towards}, T\&I2I models capture high-frequency reference features too early to preserve identity consistency. This excessive coupling prematurely locks generation. Specifically, Fig.\ref{fig3}(a) shows that $Q_{gen}$ strongly activates edge contours in $K_{ref}$ during early sampling. The spectrum in Fig.\ref{fig3}(b) further shows that these high-frequency "anchors" constrain macro-scale pose evolution. As a result, the control ability of the text query $Q_{t}$ is significantly weakened by the high-frequency shortcut, as illustrated in Fig.\ref{fig3}(c).

\textbf{Analysis for \ding{183}:} \textit{The temporal consistency via first-frame feature transport.}

In I2V modeling, the attention for the latent sequence $\mathbf{z}_{video}$ explicitly constructs cross-frame attention by facilitating direct interaction between the query vectors $\mathbf{Q}_{t,l}^i$ of subsequent frames and the key vector $\mathbf{K}_{t,l}^1$ of the first frame, as illustrated in Fig.\ref{fig3}(d). This architectural design establishes robust temporal correspondences, ensuring that visual priors from the initial frame are accurately propagated to subsequent frames, thereby maintaining identity stability while modeling temporal motion \cite{song2024save, yan2026animeagent}. Inspired by this, recent image-level methods \cite{lin2025realgeneral, chen2025unireal} adopt a "non-contiguous" frame generation architecture. By treating input and output images as a "pseudo-frame" sequence, they leverage the video model’s inherent capacity for maintaining structural and feature consistency.

\subsection{Progressive Decoupled Anchoring (PDA)}
\textbf{Motivation.} While T2I offer high compositional flexibility through noise synthesis, they struggle with subject consistency. Conversely, T\&I2I anchor identity but often fall into "high-frequency shortcuts." Current solutions typically rely on computationally expensive paradigms, such as large-scale paired data augmentation \cite{deng2025magref, xu2025withanyone} or complex feedback optimization \cite{cheng2025umo, goyal2025preventing}. We hope that an ideal T\&I2I should balance the creative freedom of T2I with the robust consistency of T\&I2I.

\textbf{Spectral Decoupling (SD).} In joint multi-modal attention, $K_{ref}$ acts as a semantic anchor. Its high-frequency components often produce sharp attention peaks that lock generation to the reference geometry. To alleviate this pose-locking effect, we reshape the reference latent $\mathcal{M}_{ref} \in \mathbb{R}^{H \times W \times D}$ and apply frequency decoupling: $\mathcal{M}_{low} = \mathcal{G}(\mathcal{M}_{ref}, \sigma), \quad \mathcal{M}_{high} = \mathcal{M}_{ref} - \mathcal{M}_{low}.$

\begin{figure*}[tbp]
\centering
\includegraphics[width=\linewidth]{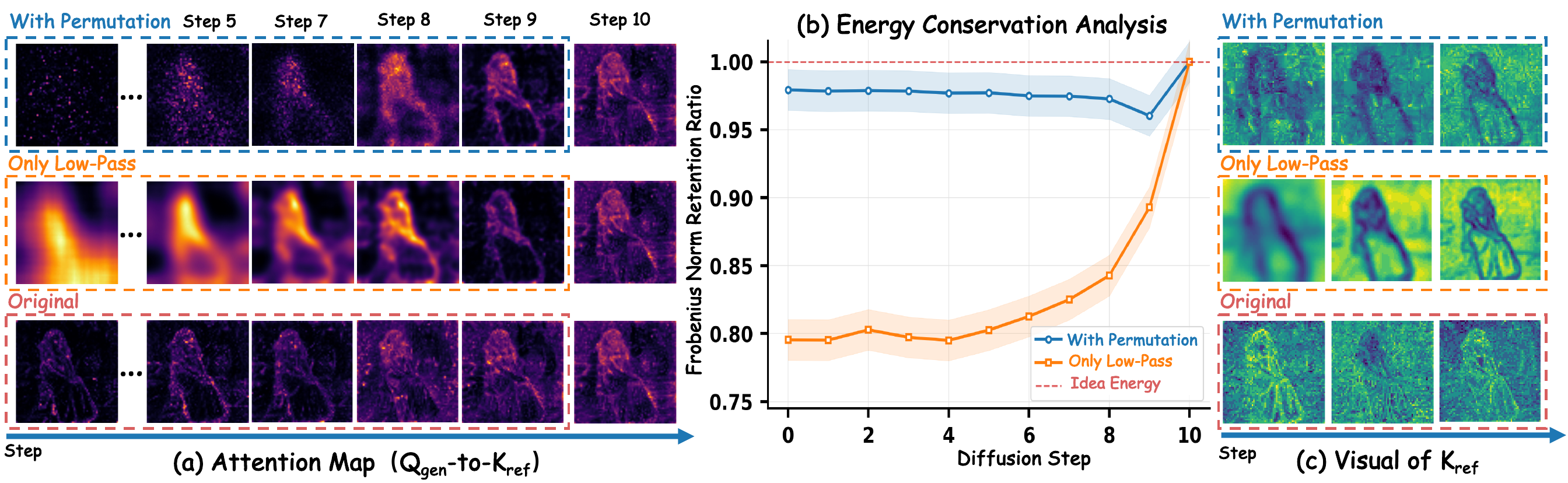}
\caption{\label{fig5} Synergistic Analysis of Spectral Decoupling and Structural Permutation.}
\vspace{-8pt}
\end{figure*}

Where $\mathcal{G}$ denotes a Gaussian low-pass filter with a kernel scale $\sigma$. While $\mathcal{M}_{low}$ effectively circumvents high-frequency structural shortcuts, it also incurs a notable loss of feature energy. As illustrated in Fig. \ref{fig5}(b), an approximately 20\% loss in \textit{Frobenius Norm} significantly degrades the feature intensity of the pre-trained model, consequently compromising the fidelity of subject reconstruction.

\textbf{Structural Permutation (SP).} To compensate for information loss from filtering while eliminating geometric bias, we apply a block-level permutation $\mathcal{P}$ to the high-frequency residual $\mathcal{M}_{high}$:
\begin{equation}
\hat{\mathcal{M}}_{ref} = \mathcal{M}_{low} + \mathcal{P}(\mathcal{M}_{high}, p),
\end{equation}
where $p$ denotes the patch size. As visualized in Fig. \ref{fig5}(c), SP redistributes high-frequency features as "disassembled components" rather than merely erasing details. By disrupting global pose correlations while preserving original statistical properties and local details, SP avoids the identity blurring and feature collapse inherent in only low-pass schemes (which retain only 0.8 of the energy).

From the perspective of attention dynamics (Fig. \ref{fig5}(a)), the smooth distribution of only low-pass features (Fig. \ref{fig5}(c)) triggers attention diffusion, which impairs precise alignment. In contrast, while SP perturbs spatial coordinates, its block structures retain local gradient transitions. his high-contrast distribution (Fig. \ref{fig5}(a)) allows the model to bypass compositional constraints while accurately extracting critical details from the "feature pool," ensuring high-fidelity reconstruction.

\textbf{Progressive Anchoring (PA).} Leveraging the coarse-to-fine nature of diffusion denoising, we introduce a time-dependent coefficient $t \in (1\to 0)$ to dynamically modulate anchoring intensity:
\begin{equation}
K_{ref}=\text{Flatten}\left( \mathcal{G}(\mathcal{M}_{ref}, \sigma(t))+\mathcal{P}(\mathcal{M}_{ref}-\mathcal{G}(\mathcal{M}_{ref}, \sigma(t)), p)\right)
\end{equation}
Macro-Evolution ($t \approx 1$): Low-pass filter suppresses high-frequency spatial anchors, forcing the model to synthesize global motion structures from the prompt $Q_{t}$ within the "spectral blind zone."

Micro-Alignment ($t \to 0$): As the global manifold stabilizes, structural signals are reintroduced as local guides for identity reconstruction without disrupting the established pose.

\subsection{Pseudo-Flow Guided Warping Feature Transport (WFT)}
\textbf{Motivation.} The superior consistency of I2V stems from feature transport under temporal coherence \cite{yan2026animeagent,burgert2025go}. Existing methods \cite{rotstein2025pathways,zhang2025image,yan2026animeagent} reformulate T2I as ``video frame prediction'' to inherit this property, but require generating redundant intermediate frames. We instead ask: \textbf{``Can T\&I2I directly simulate the `I2V generation logic' within T2I to enhance subject consistency?''}

\textbf{Formulation:} Consider a video sequence $(f_0, f_1, \dots, f_n)$ as a continuous function in 2D space, where each frame $f_j$ evolves from its predecessor $f_{j-1}$ via a mapping $T_j$. The feature vector $\Phi_j(x)$ at coordinate $x$ in the $j$-th frame is defined by the selective transport of prior features $\Phi_{j-1}$:
\begin{equation}
\Phi_j(x) = \mathcal{R} \left( \Phi_{j-1}(T_j^{-1}(x)), \Delta \phi_j \right),
\end{equation}
where $T_j$ denotes the mapping function, and $\mathcal{R}$ represents the transport operator. It propagates semantic information from the reference frame to the current frame according to the spatial mapping, with $\Delta \phi_j$ accounting for frame-specific residual updates. This formulation reveals that temporal coherence fundamentally arises from the combination of spatial feature transport and local updates.

In I2V, The 3D attention implicitly realizes the above process through Query-Key matching \cite{nam2025emergent}: $A(x, y) \propto \langle Q_j(x), K_{j-1}(y) \rangle$ which can be interpreted as a soft correspondence distribution enabling cross-frame non-rigid alignment. Functionally, it resembles a probabilistic flow field \cite{daras2024warped, song2024save}.

Motivated by this observation, we reinterpret T\&I2I architectures (e.g., FLUX.2) as a degenerate two-frame I2V process, where the reference and target images correspond to pseudo-first frame $f_0$ and pseudo-final frame $f_1$. We then construct a pseudo-flow mapping $T$ within T\&I2I to model semantic correspondence: $\Phi_1(x) = \mathcal{R}(\Phi_0(T^{-1}(x)), \Delta\phi(x))$, where pseudo-flow performs semantic transport and the residual term injects text-conditioned novel semantics.

\textbf{Method.} Unlike methods that rely on external models to establish correspondences \cite{bai2025edicho}, we compute the global pairwise similarity $S \in \mathbb{R}^{L_{gen} \times L_{ref}}$ between object $Q_{gen}$ and reference  $K_{ref}$:
\begin{equation}
S_{i, j} = \frac{\langle Q_{gen,i}, K_{ref,j} \rangle}{\|Q_{gen,i}\| \|K_{ref,j}\|}.
\end{equation}
To suppress redundant matches and enforce spatial coherence, we introduce a two-stage constraint:

\textbf{\ding{182} Local Consensus (LC).} To suppress semantic mismatches caused by repetitive textures, we introduce a neighborhood continuity assumption, i.e., correctly matched points should exhibit consistent local motion patterns. Specifically, we apply 8-neighbor shifts to $S_{i, j}$ and compute the mean of the Top-4 neighborhood responses as the local weight $W_{local}:$
\begin{equation}
W_{local} = \text{Top-K}_{mean} \left( \{ S(x+\Delta x, y+\Delta y) \mid (\Delta x, \Delta y) \in \text{Shifts} \}, k=4 \right)
\end{equation}
Using Top-4 restricts responses to valid regions near boundaries and occlusions. The refined matrix $\hat{S} = S \odot W_{local}$ effectively suppresses isolated mismatches.

\textbf{\ding{183} Mutual Nearest Neighbor (MNN).} To further improve matching robustness, we impose a bidirectional Top-$K$ constraint. A correspondence is considered valid only when target position $i$ and reference position $j$ are mutual Top-$K$ matches: $M_{valid} = M_{g \to r}(K) \cap M_{r \to g}(K)$. The optimal correspondence is then selected as $j^* = \arg\max_{j} (\hat{S}_{i,j} \cdot M_{valid})$,  which is mapped back to 2D coordinates. The offset relative to the sampling grid $\mathcal{G}$ yields the pseudo-flow field: $\mathcal{F} \in \mathbb{R}^{2 \times H \times W}$.

\textbf{Reference Feature Warping.} Given $\mathcal{F}$, the reference value $V_{ref}$ are aligned to the object feature space. Specifically, the sampling coordinates are defined as $\mathcal{G}_{sample} = \mathcal{G} + \mathcal{F}$. and bilinear resampling is performed in the normalized coordinate space $[-1,1]$, with out-of-bound regions zero-padded to avoid boundary stretching artifacts. Meanwhile, the final constraint mask is constructed from $M_{valid}$ and $\hat{S}$ to filter mismatched and occluded regions: $V_{warp} = \text{Sample}(V_{ref}, \mathcal{G}_{sample}) \odot M_{final}$.
\begin{wrapfigure}[12]{r}{0.4\textwidth} 
\centering
\vspace{-14pt} 
\includegraphics[width=\linewidth]{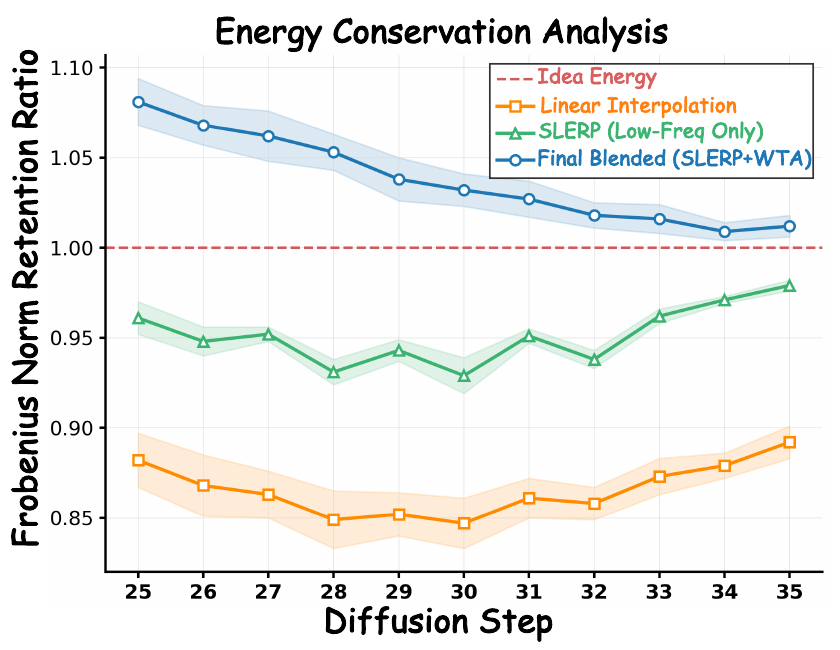}
\vspace{-14pt} 
\centering
\caption{\label{fig6}Feature Energy Analysis.}
\end{wrapfigure}
\textbf{Warped Feature Transport.} Directly injecting warped reference features $V_{warp}$ into $V_{gen}$ causes distribution mismatch. Although $V_{warp}$ preserves reference semantics, non-rigid deformation and interpolation introduce deviations from the $V_{gen}$ in local details and feature statistics. Thus, we formulate the transport process as $\mathcal{R}(V_{gen}, V_{warp})$, where $\mathcal{R}$ is a spatial frequency-decoupled operator that separately models structural consistency and detail fidelity.

\textbf{\ding{182} Low-Frequency SLERP.} Low-frequency components encode global structures such as contours and color distribution. We transport $V_{warp}^{low}$ to $V_{gen}^{low}$ using spherical linear interpolation (SLERP) \cite{alimohammadi2025cora}. \textbf{\ding{183} High-Frequency Winner-Takes-All.} High-frequency contain texture details and are sensitive to alignment errors. Direct transport may introduce ghosting or blur artifacts. Therefore, we adopt a Winner-Takes-All (WTA) strategy. The final generated value is computed as:$V_{gen} = \text{SLERP}(V_{gen}^{low}, V_{warp}^{low}) + \operatorname*{argmax}_{V \in \{V_{warp}^{high}, V_{gen}^{high}\}} \|V\|$. 

\textbf{\ding{184}Discuss.} As shown in Fig. \ref{fig6}, we analyze energy evolution using the Frobenius-norm retention ratio. Compared with linear interpolation, SLERP substantially mitigates energy attenuation, thereby better preserving the low-frequency information established in early denoising stages. Moreover, the synergy of SLERP and WTA yields a final feature energy slightly above the ideal expectation (ratio > 1.0), providing additional support for reconstructing high-frequency details in the mid-to-late denoising stages.

\section{Experiments}
\textbf{Implementation Details.}
We implement our method on FLUX.2-Klein-4B, a mainstream T\&I2I framework with a DiT-based joint self-attention architecture. Progressive spectral decoupling is applied in the first 20\% of denoising steps, while pseudo-flow-guided feature transport is injected during the intermediate phase, i.e., 50\%–70\% of steps, when semantic structures are stabilized but fine-grained identity details are still refined. All experiments are conducted at $1024 \times 1024$ resolution on NVIDIA A100 40GB GPUs. Unless otherwise specified, hyperparameters are fixed across benchmarks and prompts. Detailed hyperparameters are provided in the Appendix.

\textbf{Baselines.} We compare our method with recent SOTA SIG approaches covering both training-based and training-free paradigms: Scone \cite{wang2025scone}, UMO \cite{cheng2025umo}, UNO \cite{wu2025less}, MOSAIC \cite{she2025mosaic}, OmniGen2 \cite{wu2025omnigen2}, DreamO \cite{mou2025dreamo}, DiptychP \cite{shin2025large}, and XVerse \cite{chen2025xverse}. These methods improve subject consistency through large-scale paired training, preference optimization, explicit identity supervision, or inference-time attention/latent feature manipulation without finetuning. For fair comparison, all baselines use official checkpoints and default hyperparameters.

\textbf{Evaluation Metrics.} We evaluate our framework on the public VistoryBench \cite{zhuang2025vistorybench} and our proposed SIGBench. VistoryBench includes 1,317 cases and 344 characters. VistoryBench measures subject consistency using Character Identification Similarity (CIDS) and Character Style Similarity (CSD), prompt alignment using a multimodal large language model, aesthetic quality (AES) using Aesthetic Predictor V2.5, and diversity using Inception Score (INC) and Copy-Paste (CP) Detection metrics. On SIGBench, we use CLIP-I and DreamSim \cite{fu2023dreamsim} for subject consistency, the proposed Comprehensive Diversity Score (CDS) for diversity, CLIP-T and PA for prompt alignment, and Aesthetic Predictor V2.5 for aesthetic quality. More details are provided in the Appendix.

\begin{table*}[!t]
\caption{Quantitative Results of Various SIG Methods on ViStoryBench \cite{zhuang2025vistorybench}. CI: Character Interaction, IA: Individual Action; The \colorbox{red!40}{\strut first}, \colorbox{orange!50}{\strut second}, and \colorbox{yellow!50}{\strut third} values are highlighted.}
\centering
\setlength{\arrayrulewidth}{1.2pt}
\resizebox{\textwidth}{!}{%
\begin{tabular}{c|c|cc|cc|ccccc|c|c|c}
\hline
\multirow{2}{*}{Methods} & \multirow{2}{*}{Venue} &
\multicolumn{2}{c|}{CSD$\uparrow$} &
\multicolumn{2}{c|}{CIDS$\uparrow$} &
\multicolumn{5}{c|}{PA$\uparrow$} &
\multirow{2}{*}{AES$\uparrow$} &
\multirow{2}{*}{CP$\downarrow$} &
\multirow{2}{*}{INC$\uparrow$}\\
\cline{3-11}
& & Cross & Self & Cross & Self &
Scene & Shot & CI & IA & Avg. &
& & \\
\hline
DreamO \cite{mou2025dreamo} & SIGGA'25
& 0.489 & \pt{0.584} & 0.542 & \pt{0.643}
& 2.99 & 2.21 & 1.72 & 1.65 & 2.14
& 5.31 & 0.533 & 9.21  \\

DiptychP \cite{shin2025large} & CVPR'25
& 0.394 & 0.462 & 0.417 & 0.536
& 2.73 & 2.04 & 1.61 & 1.50 & 1.97
& 4.60 & 0.524 & 13.18  \\

XVerse \cite{chen2025xverse} & NIPS'25
& 0.229 & 0.461 & 0.342 & 0.531
& 2.96 & 2.37 & 1.89 & 1.73 & 2.23
& 4.92 & \pt{0.519} & \ps{13.67}  \\

UNO \cite{wu2025less} & ICCV'25
& \pt{0.513} & 0.576 & 0.528 & 0.632
& 3.08 & 2.52 & 2.07 & 1.91 & 2.39
& 4.94 & 0.542 & 8.27  \\

OmniGen2 \cite{wu2025omnigen2} & arXiv'25
& 0.446 & 0.546 & 0.460 & 0.587
& 3.12 & 2.46 & 1.80 & 1.64 & 2.27
& 5.19 & 0.531 & 9.87  \\

Scone \cite{wang2025scone} & CVPR'26
& 0.511 & 0.539 & 0.501 & 0.585
& 3.11 & 2.40 & 2.18 & 1.69 & 2.35
& 5.22 & \ps{0.517} & 10.60  \\

UMO \cite{cheng2025umo} & CVPR'26
& \ps{0.514} & \pf{0.635} & \pt{0.555} & \ps{0.648}
& \pt{3.25} & \pf{2.65} & 2.13 & \pt{1.93} & \pt{2.49}
& 5.05 & 0.536 & 10.35  \\

MOSAIC \cite{she2025mosaic} & ICLR'26
& 0.268 & 0.463 & 0.377 & 0.531
& 3.14 & 2.43 & \ps{2.31} & 1.96 & 2.46
& \pt{5.36} & 0.548 & \pt{13.56}  \\

FLUX.2-4B & -
& 0.489 & 0.531 & \ps{0.576} & 0.637
& \ps{3.27} & \pt{2.54} & \pt{2.27} & \ps{1.99} & \ps{2.52}
& \pf{5.42} & 0.537 & 10.97  \\

\hline
Ours(FLUX.2-4B)& -
& \pf\textbf{0.521} & \ps\textbf{{0.598}}
& \pf{\textbf{0.605}} & \pf{\textbf{0.653}}
& \pf{\textbf{3.32}} & \ps{\textbf{2.58}}
& \pf{\textbf{2.35}} & \pf\textbf{2.08}
& \pf\textbf{2.58}
& \ps{\textbf{5.39}}
& \pf{\textbf{0.516}}
& \pf{\textbf{13.88}} \\
\hline
\end{tabular}%
}
\label{tab1}
\vspace{-4pt}
\end{table*}

\begin{table*}[!t]
    \caption{Quantitative Results of Various SIG Methods on SIGBench. VRAM: Using GPU memory during inference. The \colorbox{red!40}{\strut first}, \colorbox{orange!50}{\strut second}, and \colorbox{yellow!50}{\strut third} values are highlighted.}
    \centering
    \setlength{\arrayrulewidth}{1.2pt} 
    \resizebox{\textwidth}{!}{%
    \begin{tabular}{c|c|cc|cccc|c|c|c|c|c}
    \hline
        \multirow{2}{*}{Methods} &
        \multirow{2}{*}{Venue} &
        \multicolumn{2}{c|}{DreamSim$\downarrow$} &
        \multicolumn{4}{c|}{PA$\uparrow$} &
        \multirow{2}{*}{CLIP-T$\uparrow$} &
        \multirow{2}{*}{CLIP-I$\uparrow$} &
        \multirow{2}{*}{CDS$\uparrow$} &
        \multirow{2}{*}{AES$\uparrow$} &
        \multirow{2}{*}{VRAM$\downarrow$}
        \\ \cline{3-8}
        & & GT & Ref & Scene & CI & IA & Avg. & & & & & \\ \hline
        DreamO \cite{mou2025dreamo} & SIGGA'25 & 0.273 & \pf{0.596} & 3.04 & 2.43 & 2.07 & 2.51 & \pt{0.361} & 0.834 & 0.373 & 6.00 & \pf{9.09} \\
        
        DiptychP \cite{shin2025large} & CVPR'25 & 0.459 & 0.651 & 2.72 & 2.16 & 2.45 & 2.44 & 0.332 & 0.696 & 0.339 & 5.24 & 30.01 \\
        
        XVerse \cite{chen2025xverse} & NIPS'25 & 0.339 & 0.660 & 2.95 & 2.53 & 2.12 & 2.53 & 0.358 & 0.767 & 0.366 & \pt{6.09} & 37.39 \\
  
        UNO \cite{wu2025less} & ICCV'25 & 0.315 & 0.604 & 3.21 & 2.59 & 2.28 & 2.69 & 0.345 & 0.802 & 0.366 & 5.64 & 35.09 \\
        
        OmniGen2 \cite{wu2025omnigen2} & arXiv'25 & 0.287 & 0.607 & 3.28 & 2.67 & 2.41 & 2.78 & \pf{0.363} & 0.822 & 0.380 & 5.47 & \ps{9.92} \\
        

        Scone \cite{wang2025scone} & CVPR'26 & \pt{0.261} & \ps{0.601} & \pf{3.51} & 2.88 & \pt{2.57} & \pt{2.98} & \ps{0.362} & \pt{0.838} & 0.383 & 5.83 & 30.43 \\
        
        UMO \cite{cheng2025umo} & CVPR'26 & 0.296 & \pt{0.602} & 3.38 & \pt{3.01} & 2.44 & 2.94 & 0.348 & 0.815 & 0.375 & 5.97 & 33.78 \\
        
        MOSAIC \cite{she2025mosaic} & ICLR'26 & 0.281 & 0.621 & \pt{3.43} & 2.70 & 2.56 & 2.89 & 0.354 & 0.824 & \pt{0.401} & 5.82 & 35.79 \\
        
        FLUX.2-4B  & - & \ps{0.225} & 0.643 & 3.40 & \ps{3.15} & \ps{2.63} & \ps{3.06} & 0.354 & \ps{0.847} & \ps{0.412} & \pf{6.33} & \pt{12.21} \\
        \hline
        
        Ours (FLUX.2-4B) & - & \pf\textbf{0.221} & \textbf{{0.625}} & \ps{\textbf{3.47}} & \pf{\textbf{3.34}} & \pf{\textbf{2.86}} & \pf{\textbf{3.22}} & \ps{\textbf{0.362}} & \pf{\textbf{0.854}} & \pf{\textbf{0.429}} & \ps{\textbf{6.32}} & \textbf{13.44} \\ 
        \hline
    \end{tabular}%
    }
    \label{tab2}
\vspace{-4pt}
\end{table*}

\textbf{Quantitative Results.} Table \ref{tab1} shows that our method achieves the excellent overall performance on VistoryBench. It ranks first on CSD-Cross, CIDS-Cross, CIDS-Self, and average PA, demonstrating stronger subject consistency and prompt alignment. Compared with baseline, it improves CIDS-Cross from 0.576 to 0.605 and average PA from 2.52 to 2.58, indicating that WFT improves identity-preserving feature reuse while PDA reduces reference-induced structural shortcuts. It also achieves the lowest Copy-Paste score of 0.516 and the highest Inception Score of 13.88, showing greater diversity and less reference copying.

Table \ref{tab2} reports results on SIGBench, where our method ranks first on DreamSim-GT, PA-Avg., CLIP-I, and CDS. Compared with FLUX.2-4B, it improves PA-Avg. from 3.06 to 3.22, CLIP-I from 0.847 to 0.854, and CDS from 0.412 to 0.429, showing better prompt alignment, subject consistency, and diversity. These results support our design: PDA reduces pose copying and enhances compositional flexibility, while WFT transports identity features via pseudo-flow alignment. With only minor VRAM overhead over FLUX.2-4B, our method remains more efficient than most training-based baselines and achieves superior overall performance.

\begin{figure*}[tbp]
\centering
\includegraphics[width=\linewidth]{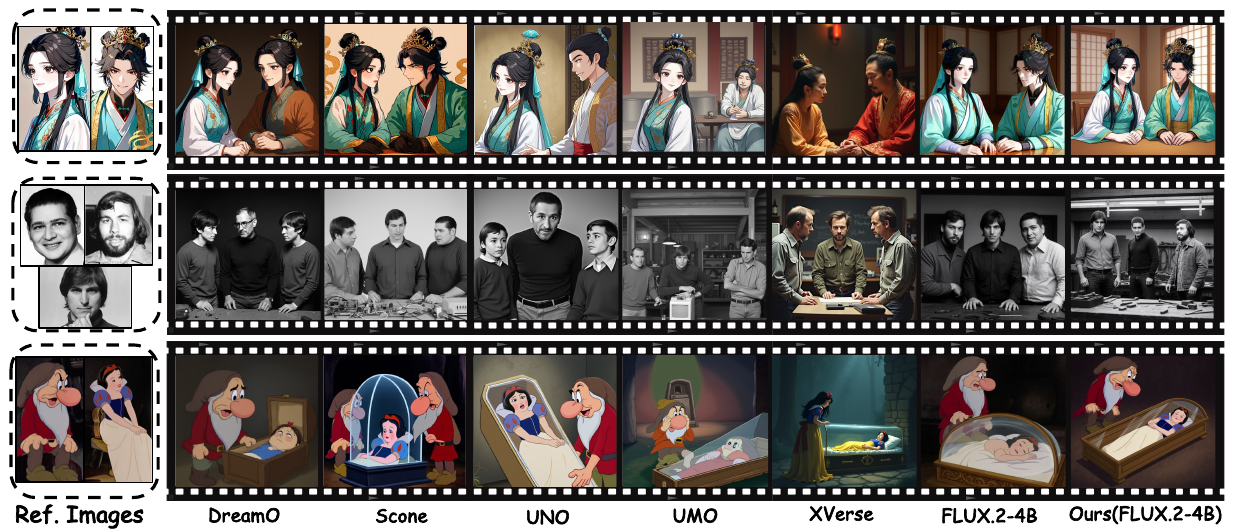}
\caption{\label{fig7}Qualitative Comparison on ViStoryBench Across Diverse Characters and Styles.}
\vspace{-4pt}
\end{figure*}

\begin{figure*}[tbp]
\centering
\includegraphics[width=\linewidth]{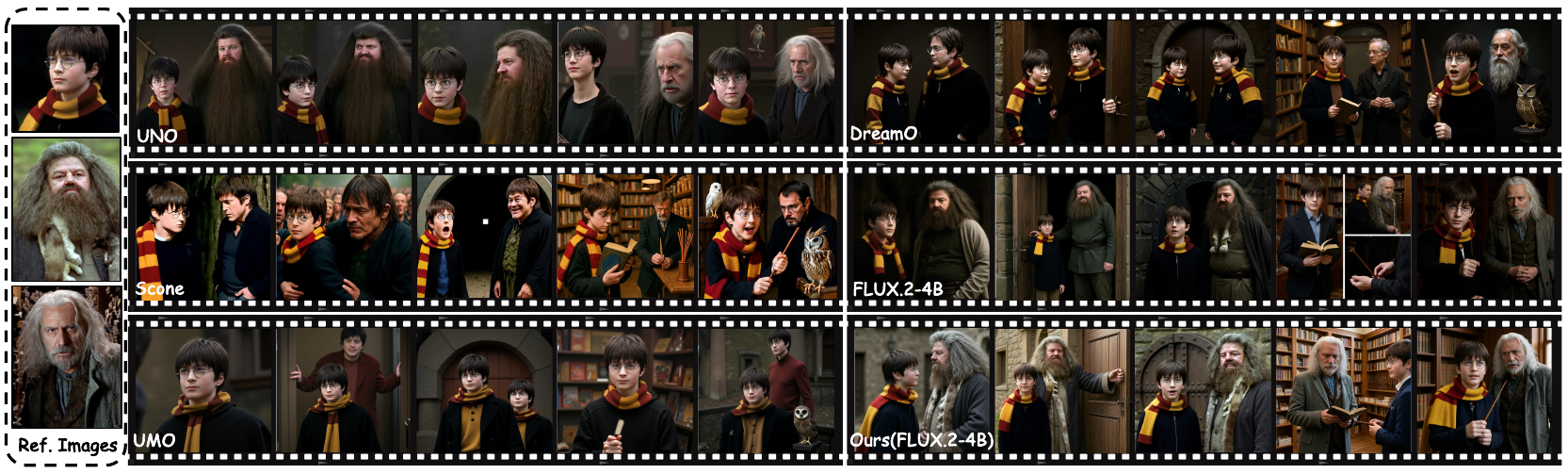}
\caption{\label{fig8}Qualitative Comparison on Multi-Character Story Generation in ViStoryBench.}
\vspace{-10pt}
\end{figure*}

\textbf{Qualitative Results.}
Fig. \ref{fig7} and \ref{fig8} show qualitative comparisons on ViStoryBench. Existing methods often exhibit identity drift, missing subjects, style inconsistency, or direct copying of the reference pose. In contrast, our method better preserves character identity and visual style while allowing prompt-driven changes in pose, interaction, and scene composition.

In multi-character cases, our method maintains more stable identities for different subjects and produces more coherent interactions. These results indicate that PDA reduces reference-structure shortcuts, while WFT restores identity-related details through feature transport.

\begin{figure*}[tbp]
\centering
\includegraphics[width=\linewidth]{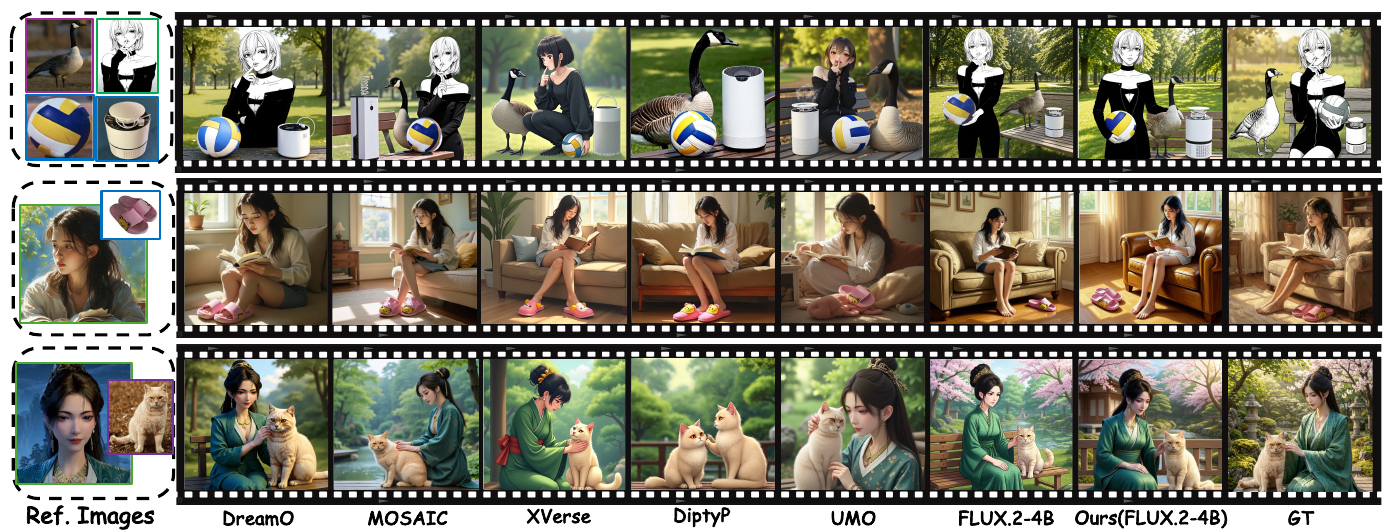}
\caption{\label{fig9} Qualitative Comparison on SIGBench with Ground Truth.}
\vspace{-4pt}
\end{figure*}

Fig. \ref{fig9} shows qualitative comparisons on SIGBench, where each case provides reference subjects and a ground-truth target image. Compared with existing methods, our method better preserves subject identity and fine-grained attributes while matching the target composition more closely. Baselines often miss objects, alter subject appearance, or over-copy the reference layout, especially in multi-subject and large-pose-change scenarios. Our results show more accurate subject-object relationships, more stable appearance preservation, and better alignment with the GT scene structure. 

\begin{table}[htbp]
\vspace{-10pt}
\centering
\caption{Human preference comparison among different methods.}
\label{tab:user_study}
\resizebox{\linewidth}{!}{
\begin{tabular}{lcccccccccc}
\toprule
\textbf{Method} & \textbf{DreamO} & \textbf{DiptychP} & \textbf{XVerse} & \textbf{UNO} & \textbf{OmniGen2} & \textbf{Scone} & \textbf{UMO} & \textbf{MOSAIC} & \textbf{FLUX.2-4B} &\textbf{Ours} \\
\midrule
\textbf{Percent (\%)$\uparrow$} & 2.79 & 1.41 & 0.62 & 10.19 & 3.03 & 12.84 & 15.95 & 4.48 & 22.75 & \textbf{26.31} \\
\bottomrule
\end{tabular}
}
\vspace{-8pt}
\end{table}

\textbf{User Study.} To complement automatic metrics, we conduct a human evaluation on 50 randomly sampled cases from the two benchmarks. For each case, all competing methods generate corresponding image sequences. We invite 20 participants, each evaluating 20 randomly selected results with the reference images and prompts. They are asked to select the preferred sequence based on subject consistency, pose diversity, and overall image quality. As shown in Table \ref{tab:user_study}, our method achieves the highest preference rate. The results further demonstrate that MorphSIG produces more consistent and visually appealing generations while maintaining stronger structural diversity. Details of the protocol and interface are provided in the Appendix.

\begin{wraptable}{r}{0.4\linewidth}
    \vspace{-0.8em}
    \centering
    \setlength{\abovecaptionskip}{0pt}
    \setlength{\belowcaptionskip}{2pt}
    \captionsetup{width=\linewidth}
    \caption{The ablation study of PDA.}
    \label{tab:x1}

    \resizebox{\linewidth}{!}{%
        \setlength{\tabcolsep}{1.8pt}
        \renewcommand{\arraystretch}{0.98}
        \begin{tabular}{@{}lcccc@{}}
            \toprule[1pt]
            & CIDS$\uparrow$& INC$\uparrow$ & PA$\uparrow$ & AES$\uparrow$ \\
            \midrule
            w/o PDA & 0.614 & 13.41 & 2.37 & 5.30 \\    
            w/o SP  & 0.556 & 13.64 & \textbf{2.72}  & 5.33\\     
            w/o SD  & 0.618 & 13.57 & 2.39 & 5.36\\    
            \midrule
            High-pass only & 0.485 & 12.81 & 1.92 & 5.05 \\   
            \midrule
            Full PDA & \textbf{0.629} & \textbf{13.88} & 2.58 & \textbf{5.39} \\ 
            \bottomrule[1pt]
        \end{tabular}%
    }
    \vspace{-0.8em}
\end{wraptable}

\textbf{Ablation Study.} Tables \ref{tab:x1} and \ref{tab:x2} report the ablation results of PDA and WFT. Removing PDA degrades CIDS, INC, and PA, indicating excessive reliance on reference structures and reduced flexibility in text-driven pose and layout generation. Specifically, removing SP weakens identity-preserving details, while removing SD reduces pose diversity, demonstrating their respective roles in identity retention and structural decoupling. Removing WFT causes a significant drop in CIDS, highlighting the importance of pseudo-flow-guided feature transport for maintaining subject consistency under large structural variations. Further removing MNN, LF-SLERP, or HF-WTA consistently lowers performance, showing that bidirectional matching, low-frequency fusion, and high-frequency detail selection jointly improve structural stability and texture fidelity. 

\begin{wraptable}{r}{0.4\linewidth}
    \vspace{-0.8em}
    \centering
    \setlength{\abovecaptionskip}{0pt}
    \setlength{\belowcaptionskip}{2pt}
    \captionsetup{width=\linewidth}
    \caption{The ablation study of WFT.}
    \label{tab:x2}
    \resizebox{\linewidth}{!}{%
        \setlength{\tabcolsep}{1.8pt}
        \renewcommand{\arraystretch}{0.98}
        \begin{tabular}{@{}lcccc@{}}
            \toprule[1pt]
         & CIDS$\uparrow$& INC$\uparrow$ & PA$\uparrow$ & AES$\uparrow$ \\
            \midrule
             w/o WFT & 0.541 & 13.47 &  \textbf{2.64} & 5.25   \\ 
             w/o MNN  & 0.578 & 13.32 &  2.47 & 5.18   \\ 
             \midrule
             w/o HF-WTA & 0.601 & 13.65 & 2.52  & 5.14   \\ 
             w/o LF-SLERP & 0.584 & 13.30 & 2.34 & 5.19  \\ 
             ADD & 0.561 & 12.93 & 2.16 & 5.05   \\ 
            \midrule
            Full WFT & \textbf{0.629} & \textbf{13.88} & 2.58 & \textbf{5.39} \\ 
            \bottomrule[1pt]
        \end{tabular}%
    }
    \vspace{-0.8em}
\end{wraptable}

\begin{figure*}[tbp]
\centering
\includegraphics[width=\linewidth]{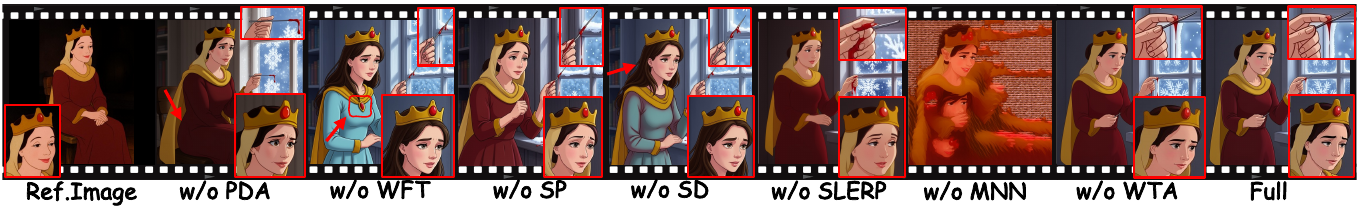}
\caption{\label{fig10} Qualitative Comparison of Ablation Study.}
\vspace{-2pt}
\end{figure*}

Fig. \ref{fig10} illustrates the visual differences under various ablation settings. Without PDA, the generated results tend to replicate the original pose, limiting the controllability of text-driven generation. Removing SP or SD further causes identity-detail degradation and insufficient structural disentanglement, respectively. For WFT-related ablations, removing WFT or MNN introduces local misalignment and identity drift, while removing LF-SLERP or HF-WTA results in unstable structures or blurred textures. In contrast, the full model better preserves subject consistency while generating novel poses and scenes aligned with the text prompt, demonstrating the complementary effects of PDA and WFT in harmonizing identity retention with generative diversity.

\begin{figure}[tbp]
\centering
\includegraphics[width=\linewidth]{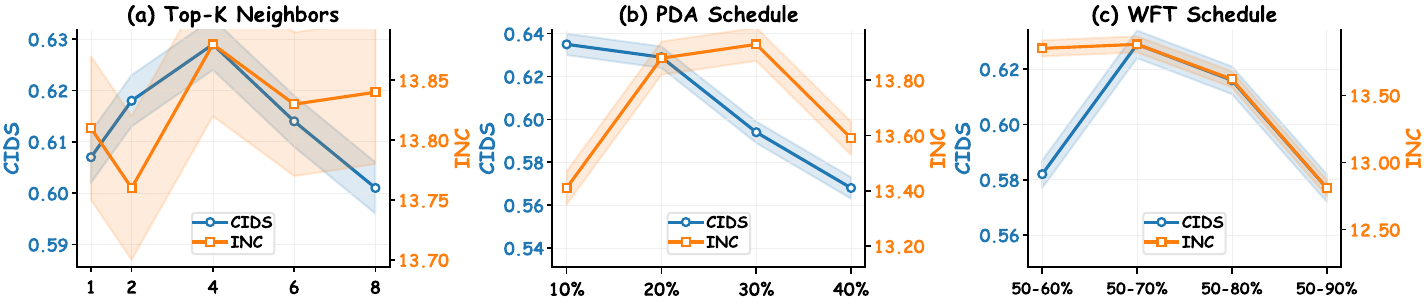} 
\vspace{-4pt}
\caption{\textbf{Ablation analysis.} (a) Effect of neighborhood size $K$ in Local Consensus. (b, c) Sensitivity of PDA and WFT scheduling ranges.}
\label{fig:ablation_sensitivity}
\vspace{-4pt}
\end{figure}

\noindent\textbf{Local Consensus Analysis.} As shown in Fig.~\ref{fig:ablation_sensitivity}(a), selecting $K=4$ in Local Consensus achieves the best balance between identity fidelity (CIDS) and diversity (INC). A smaller $K$ admits spurious matches from texture ambiguity, whereas a larger $K$ over-suppresses valid correspondences by incorporating noisy boundary neighbors.

\noindent\textbf{Scheduling Sensitivity Analysis.} 
As shown in Fig.~\ref{fig:ablation_sensitivity}(b, c), our default schedule achieves the optimal balance between identity fidelity and generation diversity. Terminating PDA prematurely (10\%) locks macro-composition via high-frequency shortcuts, whereas prolonging it ($>$20\%) impairs fine-grained detail alignment. Similarly, WFT requires sufficient propagation without over-intervention: truncating it early ($<$70\%) hinders detail consistency, while extending it into late denoising ($>$80\%) introduces distribution mismatch and artifacts. Thus, our setting (PDA: 20\%, WFT: 50\%--70\%) reliably optimizes performance.

\begin{table}[htbp]
\vspace{-4pt}
\centering
\caption{Comparison of computational cost on a single NVIDIA A100 GPU.}
\label{tab:cost}
\resizebox{\linewidth}{!}{
\begin{tabular}{lcccccccccc}
\toprule
\textbf{Method} & \textbf{DreamO} & \textbf{DiptychP} & \textbf{XVerse} & \textbf{UNO} & \textbf{OmniGen2} & \textbf{Scone} & \textbf{UMO} & \textbf{MOSAIC} & \textbf{FLUX.2-4B} & \textbf{Ours} \\
\midrule
\textbf{Training-free} & $\times$ & \checkmark & $\times$ & $\times$ & $\times$ & $\times$ & $\times$ & $\times$ & -- & \textbf{\checkmark} \\
\textbf{Latency (s)} & 76.9 & 169.4 & 126.2 & 82.6 & 171.7 & 98.5 & 84.3 & 154.7 & 115.6 & \textbf{130.1} \\
\textbf{VRAM (GB)} & 9.09 & 30.01 & 37.39 & 35.09 & 9.92 & 30.43 & 33.78 & 35.79 & 12.21 & \textbf{13.44} \\
\bottomrule
\end{tabular}
}
\end{table}

\noindent\textbf{Computational Cost.} As shown in Table \ref{tab:cost}, MorphSIG is training-free method, incurring zero training cost. Compared to the native FLUX.2-4B (115.6s / 12.21GB), it introduces only marginal overhead in inference latency (130.1s, +12.5\%) and peak VRAM (13.44GB, +10.1\%), while remaining substantially lighter than training-based baselines exceeding 30GB (e.g., XVerse and MOSAIC).

\section{Conclusion}
We presented MorphSIG, a training-free framework for subject-driven image generation. By identifying high-frequency reference shortcuts as a major cause of copy-paste artifacts, we proposed Progressive Decoupled Anchoring to relax structural locking while preserving identity-related statistics. We further reformulated T\&I2I as pseudo-video generation and introduced Pseudo-flow Guided Feature Transport to improve subject consistency through latent-space feature transport without generating intermediate frames. In addition, we built a new SIG benchmark with diverse styles and high-quality ground truth. Experiments on public and proposed benchmarks show that MorphSIG achieves a strong balance between identity fidelity and generation diversity. These results demonstrate the effectiveness of frequency-decoupled conditioning and pseudo-flow-guided transport for controllable and consistent subject-driven generation.

\vfill\pagebreak
\medskip
{
	\small
	\bibliographystyle{plain}
	\bibliography{reference}

@inproceedings{ruiz2023dreambooth,
  title={Dreambooth: Fine tuning text-to-image diffusion models for subject-driven generation},
  author={Ruiz, Nataniel and Li, Yuanzhen and Jampani, Varun and Pritch, Yael and Rubinstein, Michael and Aberman, Kfir},
  booktitle={Proceedings of the IEEE/CVF Conference on Computer Vision and Pattern Recognition},
  pages={22500--22510},
  year={2023}
}

@inproceedings{kumari2023multi,
  title={Multi-concept customization of text-to-image diffusion},
  author={Kumari, Nupur and Zhang, Bingliang and Zhang, Richard and Shechtman, Eli and Zhu, Jun-Yan},
  booktitle={Proceedings of the IEEE/CVF Conference on Computer Vision and Pattern Recognition},
  pages={1931--1941},
  year={2023}
}

@article{gu2023mix,
  title={Mix-of-show: Decentralized low-rank adaptation for multi-concept customization of diffusion models},
  author={Gu, Yuchao and Wang, Xintao and Wu, Jay Zhangjie and Shi, Yujun and Chen, Yunpeng and Fan, Zihan and Xiao, Wuyou and Zhao, Rui and Chang, Shuning and Wu, Weijia and others},
  journal={Advances in Neural Information Processing Systems},
  volume={36},
  pages={15890--15902},
  year={2023}
}

@inproceedings{mou2024t2i,
  title={T2i-adapter: Learning adapters to dig out more controllable ability for text-to-image diffusion models},
  author={Mou, Chong and Wang, Xintao and Xie, Liangbin and Wu, Yanze and Zhang, Jian and Qi, Zhongang and Shan, Ying},
  booktitle={Proceedings of the AAAI conference on artificial intelligence},
  volume={38},
  number={5},
  pages={4296--4304},
  year={2024}
}

@inproceedings{zhang2025ip,
  title={IP-Prompter: Training-Free Theme-Specific Image Generation via Dynamic Visual Prompting},
  author={Zhang, Yuxin and Luo, Minyan and Dong, Weiming and Yang, Xiao and Huang, Haibin and Ma, Chongyang and Deussen, Oliver and Lee, Tong-Yee and Xu, Changsheng},
  booktitle={Proceedings of the Special Interest Group on Computer Graphics and Interactive Techniques Conference Conference Papers},
  pages={1--12},
  year={2025}
}

@article{ye2023ip,
  title={Ip-adapter: Text compatible image prompt adapter for text-to-image diffusion models},
  author={Ye, Hu and Zhang, Jun and Liu, Sibo and Han, Xiao and Yang, Wei},
  journal={arXiv preprint arXiv:2308.06721},
  year={2023}
}

@inproceedings{ding2024freecustom,
  title={Freecustom: Tuning-free customized image generation for multi-concept composition},
  author={Ding, Ganggui and Zhao, Canyu and Wang, Wen and Yang, Zhen and Liu, Zide and Chen, Hao and Shen, Chunhua},
  booktitle={Proceedings of the IEEE/CVF Conference on Computer Vision and Pattern Recognition},
  pages={9089--9098},
  year={2024}
}

@article{tewel2024training,
  title={Training-free consistent text-to-image generation},
  author={Tewel, Yoad and Kaduri, Omri and Gal, Rinon and Kasten, Yoni and Wolf, Lior and Chechik, Gal and Atzmon, Yuval},
  journal={ACM Transactions on Graphics (TOG)},
  volume={43},
  number={4},
  pages={1--18},
  year={2024},
  publisher={ACM New York, NY, USA}
}

@article{mazuz2025consistyle,
  title={ConsiStyle: Style Diversity in Training-Free Consistent T2I Generation},
  author={Mazuz, Yohai and Bruner, Janna and Wolf, Lior},
  journal={ACM Transactions on Graphics (TOG)},
  volume={44},
  number={6},
  pages={1--16},
  year={2025},
  publisher={ACM New York, NY, USA}
}

@inproceedings{wang2025characonsist,
  title={Characonsist: Fine-grained consistent character generation},
  author={Wang, Mengyu and Ding, Henghui and Peng, Jianing and Zhao, Yao and Chen, Yunpeng and Wei, Yunchao},
  booktitle={Proceedings of the IEEE/CVF International Conference on Computer Vision},
  pages={16058--16067},
  year={2025}
}

@inproceedings{li2025consistent,
  title={Consistent Story Generation: Unlocking the Potential of Zigzag Sampling},
  author={Li, Mingxiao and Ning, Mang and Moens, Marie-Francine},
  booktitle={The Thirty-ninth Annual Conference on Neural Information Processing Systems},
  year={2025}
}

@inproceedings{wu2025less,
  title={Less-to-more generalization: Unlocking more controllability by in-context generation},
  author={Wu, Shaojin and Huang, Mengqi and Wu, Wenxu and Cheng, Yufeng and Ding, Fei and He, Qian},
  booktitle={Proceedings of the IEEE/CVF International Conference on Computer Vision},
  pages={18682--18692},
  year={2025}
}

@article{cheng2025umo,
  title={UMO: Scaling Multi-Identity Consistency for Image Customization via Matching Reward},
  author={Cheng, Yufeng and Wu, Wenxu and Wu, Shaojin and Huang, Mengqi and Ding, Fei and He, Qian},
  journal={arXiv preprint arXiv:2509.06818},
  year={2025}
}

@article{wu2025multicrafter,
  title={MultiCrafter: High-Fidelity Multi-Subject Generation via Disentangled Attention and Identity-Aware Preference Alignment},
  author={Wu, Tao and Jiang, Yibo and Lu, Yehao and Wang, Zhizhong and Huang, Zeyi and Qin, Zequn and Li, Xi},
  journal={arXiv preprint arXiv:2509.21953},
  year={2025}
}

@inproceedings{mou2025dreamo,
  title={Dreamo: A unified framework for image customization},
  author={Mou, Chong and Wu, Yanze and Wu, Wenxu and Guo, Zinan and Zhang, Pengze and Cheng, Yufeng and Luo, Yiming and Ding, Fei and Zhang, Shiwen and Li, Xinghui and others},
  booktitle={Proceedings of the SIGGRAPH Asia 2025 Conference Papers},
  pages={1--12},
  year={2025}
}

@article{wu2025uso,
  title={Uso: Unified style and subject-driven generation via disentangled and reward learning},
  author={Wu, Shaojin and Huang, Mengqi and Cheng, Yufeng and Wu, Wenxu and Tian, Jiahe and Luo, Yiming and Ding, Fei and He, Qian},
  journal={arXiv preprint arXiv:2508.18966},
  year={2025}
}

@article{wu2025omnigen2,
  title={Omnigen2: Exploration to advanced multimodal generation},
  author={Wu, Chenyuan and Zheng, Pengfei and Yan, Ruiran and Xiao, Shitao and Luo, Xin and Wang, Yueze and Li, Wanli and Jiang, Xiyan and Liu, Yexin and Zhou, Junjie and others},
  journal={arXiv preprint arXiv:2506.18871},
  year={2025}
}

@inproceedings{xiao2025omnigen,
  title={Omnigen: Unified image generation},
  author={Xiao, Shitao and Wang, Yueze and Zhou, Junjie and Yuan, Huaying and Xing, Xingrun and Yan, Ruiran and Li, Chaofan and Wang, Shuting and Huang, Tiejun and Liu, Zheng},
  booktitle={Proceedings of the IEEE/CVF Conference on Computer Vision and Pattern Recognition},
  pages={13294--13304},
  year={2025}
}

@inproceedings{hu2025dynamicid,
  title={Dynamicid: Zero-shot multi-id image personalization with flexible facial editability},
  author={Hu, Xirui and Wang, Jiahao and Chen, Hao and Zhang, Weizhan and Wang, Benqi and Li, Yikun and Nan, Haishun},
  booktitle={Proceedings of the IEEE/CVF International Conference on Computer Vision},
  pages={10549--10559},
  year={2025}
}

@article{she2025mosaic,
  title={MOSAIC: Multi-Subject Personalized Generation via Correspondence-Aware Alignment and Disentanglement},
  author={She, Dong and Fu, Siming and Liu, Mushui and Jin, Qiaoqiao and Wang, Hualiang and Liu, Mu and Jiang, Jidong},
  journal={arXiv preprint arXiv:2509.01977},
  year={2025}
}

@article{dalva2025lorashop,
  title={Lorashop: Training-free multi-concept image generation and editing with rectified flow transformers},
  author={Dalva, Yusuf and Yesiltepe, Hidir and Yanardag, Pinar},
  journal={arXiv preprint arXiv:2505.23758},
  year={2025}
}

@inproceedings{tan2025ominicontrol,
  title={Ominicontrol: Minimal and universal control for diffusion transformer},
  author={Tan, Zhenxiong and Liu, Songhua and Yang, Xingyi and Xue, Qiaochu and Wang, Xinchao},
  booktitle={Proceedings of the IEEE/CVF International Conference on Computer Vision},
  pages={14940--14950},
  year={2025}
}

@inproceedings{zhang2025freecus,
  title={FreeCus: Free Lunch Subject-driven Customization in Diffusion Transformers},
  author={Zhang, Yanbing and Wang, Zhe and Zhou, Qin and Yang, Mengping},
  booktitle={Proceedings of the IEEE/CVF International Conference on Computer Vision},
  pages={15521--15531},
  year={2025}
}

@article{hu2026storytailor,
  title={StoryTailor: A Zero-Shot Pipeline for Action-Rich Multi-Subject Visual Narratives},
  author={Hu, Jinghao and Zhang, Yuhe and Geng, GuoHua and Li, Kang and Zhang, Han},
  journal={arXiv preprint arXiv:2602.21273},
  year={2026}
}

@inproceedings{chen2026lamic,
  title={Lamic: Layout-aware multi-image composition via scalability of multimodal diffusion transformer},
  author={Chen, Yuzhuo and Ma, Zehua and Wang, Jianhua and Kang, Kai and Yao, Shunyu and Zhang, Weiming},
  booktitle={Proceedings of the AAAI Conference on Artificial Intelligence},
  volume={40},
  number={4},
  pages={3174--3182},
  year={2026}
}

@article{kang2025flux,
  title={Flux Already Knows--Activating Subject-Driven Image Generation without Training},
  author={Kang, Hao and Fotiadis, Stathi and Jiang, Liming and Yan, Qing and Jia, Yumin and Liu, Zichuan and Chong, Min Jin and Lu, Xin},
  journal={arXiv preprint arXiv:2504.11478},
  year={2025}
}

@article{blattmann2023stable,
  title={Stable video diffusion: Scaling latent video diffusion models to large datasets},
  author={Blattmann, Andreas and Dockhorn, Tim and Kulal, Sumith and Mendelevitch, Daniel and Kilian, Maciej and Lorenz, Dominik and Levi, Yam and English, Zion and Voleti, Vikram and Letts, Adam and others},
  journal={arXiv preprint arXiv:2311.15127},
  year={2023}
}

@inproceedings{blattmann2023align,
  title={Align your latents: High-resolution video synthesis with latent diffusion models},
  author={Blattmann, Andreas and Rombach, Robin and Ling, Huan and Dockhorn, Tim and Kim, Seung Wook and Fidler, Sanja and Kreis, Karsten},
  booktitle={Proceedings of the IEEE/CVF Conference on Computer Vision and Pattern Recognition},
  pages={22563--22575},
  year={2023}
}

@article{chen2023videocrafter1,
  title={Videocrafter1: Open diffusion models for high-quality video generation},
  author={Chen, Haoxin and Xia, Menghan and He, Yingqing and Zhang, Yong and Cun, Xiaodong and Yang, Shaoshu and Xing, Jinbo and Liu, Yaofang and Chen, Qifeng and Wang, Xintao and others},
  journal={arXiv preprint arXiv:2310.19512},
  year={2023}
}

@article{guo2023animatediff,
  title={Animatediff: Animate your personalized text-to-image diffusion models without specific tuning},
  author={Guo, Yuwei and Yang, Ceyuan and Rao, Anyi and Liang, Zhengyang and Wang, Yaohui and Qiao, Yu and Agrawala, Maneesh and Lin, Dahua and Dai, Bo},
  journal={arXiv preprint arXiv:2307.04725},
  year={2023}
}

@article{fu2025imontage,
  title={iMontage: Unified, Versatile, Highly Dynamic Many-to-many Image Generation},
  author={Fu, Zhoujie and Zeng, Xianfang and Lan, Jinghong and Liao, Xinyao and Chen, Cheng and Chen, Junyi and Wei, Jiacheng and Cheng, Wei and Liu, Shiyu and Chen, Yunuo and others},
  journal={arXiv preprint arXiv:2511.20635},
  year={2025}
}

@inproceedings{rotstein2025pathways,
  title={Pathways on the image manifold: Image editing via video generation},
  author={Rotstein, Noam and Yona, Gal and Silver, Daniel and Velich, Roy and Bensa{\"\i}d, David and Kimmel, Ron},
  booktitle={Proceedings of the IEEE/CVF Conference on Computer Vision and Pattern Recognition},
  pages={7857--7866},
  year={2025}
}

@article{tarres2026placid,
  title={PLACID: Identity-Preserving Multi-Object Compositing via Video Diffusion with Synthetic Trajectories},
  author={Tarr{\'e}s, Gemma Canet and Baradad, Manel and Moreno-Noguer, Francesc and Li, Yumeng},
  journal={arXiv preprint arXiv:2602.00267},
  year={2026}
}

@inproceedings{chen2025unireal,
  title={Unireal: Universal image generation and editing via learning real-world dynamics},
  author={Chen, Xi and Zhang, Zhifei and Zhang, He and Zhou, Yuqian and Kim, Soo Ye and Liu, Qing and Li, Yijun and Zhang, Jianming and Zhao, Nanxuan and Wang, Yilin and others},
  booktitle={Proceedings of the IEEE/CVF Conference on Computer Vision and Pattern Recognition},
  pages={12501--12511},
  year={2025}
}

@article{tong2026cof,
  title={CoF-T2I: Video Models as Pure Visual Reasoners for Text-to-Image Generation},
  author={Tong, Chengzhuo and Chang, Mingkun and Zhang, Shenglong and Wang, Yuran and Liang, Cheng and Zhao, Zhizheng and An, Ruichuan and Zeng, Bohan and Shi, Yang and Dai, Yifan and others},
  journal={arXiv preprint arXiv:2601.10061},
  year={2026}
}

@article{zhang2025image,
  title={Are Image-to-Video Models Good Zero-Shot Image Editors?},
  author={Zhang, Zechuan and Chen, Zhenyuan and Yang, Zongxin and Yang, Yi},
  journal={arXiv preprint arXiv:2511.19435},
  year={2025}
}

@inproceedings{lin2025realgeneral,
  title={Realgeneral: Unifying visual generation via temporal in-context learning with video models},
  author={Lin, Yijing and Huang, Mengqi and Zhuang, Shuhan and Mao, Zhendong},
  booktitle={Proceedings of the IEEE/CVF International Conference on Computer Vision},
  pages={14994--15004},
  year={2025}
}

@article{wu2025chronoedit,
  title={Chronoedit: Towards temporal reasoning for image editing and world simulation},
  author={Wu, Jay Zhangjie and Ren, Xuanchi and Shen, Tianchang and Cao, Tianshi and He, Kai and Lu, Yifan and Gao, Ruiyuan and Xie, Enze and Lan, Shiyi and Alvarez, Jose M and others},
  journal={arXiv preprint arXiv:2510.04290},
  year={2025}
}

@article{yan2026animeagent,
  title={Animeagent: Is the multi-agent via image-to-video models a good disney storytelling artist?},
  author={Yan, Hailong and Liu, Shice and Wang, Tao and Zhang, Xiangtao and Zhong, Yijie and Chen, Jinwei and Zhang, Le and Li, Bo},
  journal={arXiv preprint arXiv:2602.20664},
  year={2026}
}

@article{nam2025emergent,
  title={Emergent temporal correspondences from video diffusion transformers},
  author={Nam, Jisu and Son, Soowon and Chung, Dahyun and Kim, Jiyoung and Jin, Siyoon and Hur, Junhwa and Kim, Seungryong},
  journal={arXiv preprint arXiv:2506.17220},
  year={2025}
}

@article{yi2024towards,
  title={Towards understanding the working mechanism of text-to-image diffusion model},
  author={Yi, Mingyang and Li, Aoxue and Xin, Yi and Li, Zhenguo},
  journal={Advances in Neural Information Processing Systems},
  volume={37},
  pages={55342--55369},
  year={2024}
}

@article{goyal2025preventing,
  title={Preventing shortcuts in adapter training via providing the shortcuts},
  author={Goyal, Anujraaj Argo and Qian, Guocheng Gordon and Coskun, Huseyin and Gupta, Aarush and Tam, Himmy and Ostashev, Daniil and Hu, Ju and Sagar, Dhritiman and Tulyakov, Sergey and Aberman, Kfir and others},
  journal={arXiv preprint arXiv:2510.20887},
  year={2025}
}

@article{li2026disentangling,
  title={Disentangling to Re-couple: Resolving the Similarity-Controllability Paradox in Subject-Driven Text-to-Image Generation},
  author={Li, Shuang and Deng, Chao and Chen, Hang and Liu, Liqun and Hu, Zhenyu and Cao, Te and Xue, Mengge and Chen, Yuan and Shu, Peng and Yu, Huan and others},
  journal={arXiv preprint arXiv:2604.00849},
  year={2026}
}

@article{xu2025withanyone,
  title={Withanyone: Towards controllable and id consistent image generation},
  author={Xu, Hengyuan and Cheng, Wei and Xing, Peng and Fang, Yixiao and Wu, Shuhan and Wang, Rui and Zeng, Xianfang and Jiang, Daxin and Yu, Gang and Ma, Xingjun and others},
  journal={arXiv preprint arXiv:2510.14975},
  year={2025}
}

@article{deng2025magref,
  title={MAGREF: Masked Guidance for Any-Reference Video Generation with Subject Disentanglement},
  author={Deng, Yufan and Yin, Yuanyang and Guo, Xun and Wang, Yizhi and Fang, Jacob Zhiyuan and Yuan, Shenghai and Yang, Yiding and Wang, Angtian and Liu, Bo and Huang, Haibin and others},
  journal={arXiv preprint arXiv:2505.23742},
  year={2025}
}

@inproceedings{song2024save,
  title={SAVE: Protagonist Diversification with S tructure A gnostic V ideo E diting},
  author={Song, Yeji and Shin, Wonsik and Lee, Junsoo and Kim, Jeesoo and Kwak, Nojun},
  booktitle={European Conference on Computer Vision},
  pages={41--57},
  year={2024},
  organization={Springer}
}

@inproceedings{bai2025edicho,
  title={Edicho: Consistent image editing in the wild},
  author={Bai, Qingyan and Ouyang, Hao and Xu, Yinghao and Wang, Qiuyu and Yang, Ceyuan and Cheng, Ka Leong and Shen, Yujun and Chen, Qifeng},
  booktitle={Proceedings of the IEEE/CVF International Conference on Computer Vision},
  pages={15277--15287},
  year={2025}
}

@article{daras2024warped,
  title={Warped diffusion: Solving video inverse problems with image diffusion models},
  author={Daras, Giannis and Nie, Weili and Kreis, Karsten and Dimakis, Alexandros G and Mardani, Morteza and Kovachki, Nikola B and Vahdat, Arash},
  journal={Advances in Neural Information Processing Systems},
  volume={37},
  pages={101116--101143},
  year={2024}
}

@inproceedings{burgert2025go,
  title={Go-with-the-flow: Motion-controllable video diffusion models using real-time warped noise},
  author={Burgert, Ryan and Xu, Yuancheng and Xian, Wenqi and Pilarski, Oliver and Clausen, Pascal and He, Mingming and Ma, Li and Deng, Yitong and Li, Lingxiao and Mousavi, Mohsen and others},
  booktitle={Proceedings of the Computer Vision and Pattern Recognition Conference},
  pages={13--23},
  year={2025}
}

@inproceedings{alimohammadi2025cora,
  title={Cora: Correspondence-aware image editing using few step diffusion},
  author={Alimohammadi, Amirhossein and Mikaeili, Aryan and Nag, Sauradip and Hassanpour, Negar and Tagliasacchi, Andrea and Mahdavi-Amiri, Ali},
  booktitle={Proceedings of the Special Interest Group on Computer Graphics and Interactive Techniques Conference Conference Papers},
  pages={1--11},
  year={2025}
}

@article{zhuang2025vistorybench,
  title={Vistorybench: Comprehensive benchmark suite for story visualization},
  author={Zhuang, Cailin and Huang, Ailin and Hu, Yaoqi and Wu, Jingwei and Cheng, Wei and Liao, Jiaqi and Wang, Hongyuan and Liao, Xinyao and Cai, Weiwei and Xu, Hengyuan and others},
  journal={arXiv preprint arXiv:2505.24862},
  year={2025}
}

@inproceedings{feng2026personalize,
  title={Personalize anything for free with diffusion transformer},
  author={Feng, Haoran and Huang, Zehuan and Li, Lin and Sheng, Lu},
  booktitle={Proceedings of the AAAI Conference on Artificial Intelligence},
  volume={40},
  number={5},
  pages={3921--3929},
  year={2026}
}

@inproceedings{chan2024improving,
  title={Improving subject-driven image synthesis with subject-agnostic guidance},
  author={Chan, Kelvin CK and Zhao, Yang and Jia, Xuhui and Yang, Ming-Hsuan and Wang, Huisheng},
  booktitle={Proceedings of the IEEE/CVF Conference on Computer Vision and Pattern Recognition},
  pages={6733--6742},
  year={2024}
}

@article{chen2025xverse,
  title={Xverse: Consistent multi-subject control of identity and semantic attributes via dit modulation},
  author={Chen, Bowen and Zhao, Mengyi and Sun, Haomiao and Chen, Li and Wang, Xu and Du, Kang and Wu, Xinglong},
  journal={arXiv preprint arXiv:2506.21416},
  year={2025}
}

@inproceedings{gao2025postermaker,
  title={Postermaker: Towards high-quality product poster generation with accurate text rendering},
  author={Gao, Yifan and Lin, Zihang and Liu, Chuanbin and Zhou, Min and Ge, Tiezheng and Zheng, Bo and Xie, Hongtao},
  booktitle={Proceedings of the Computer Vision and Pattern Recognition Conference},
  pages={8083--8093},
  year={2025}
}

@article{wang2024ms,
  title={Ms-diffusion: Multi-subject zero-shot image personalization with layout guidance},
  author={Wang, Xierui and Fu, Siming and Huang, Qihan and He, Wanggui and Jiang, Hao},
  journal={arXiv preprint arXiv:2406.07209},
  year={2024}
}

@inproceedings{shin2025large,
  title={Large-scale text-to-image model with inpainting is a zero-shot subject-driven image generator},
  author={Shin, Chaehun and Choi, Jooyoung and Kim, Heeseung and Yoon, Sungroh},
  booktitle={Proceedings of the Computer Vision and Pattern Recognition Conference},
  pages={7986--7996},
  year={2025}
}

@article{wang2025scone,
  title={Scone: Bridging Composition and Distinction in Subject-Driven Image Generation via Unified Understanding-Generation Modeling},
  author={Wang, Yuran and Zeng, Bohan and Tong, Chengzhuo and Liu, Wenxuan and Shi, Yang and Ma, Xiaochen and Liang, Hao and Zhang, Yuanxing and Zhang, Wentao},
  journal={arXiv preprint arXiv:2512.12675},
  year={2025}
}

@article{fu2023dreamsim,
  title={Dreamsim: Learning new dimensions of human visual similarity using synthetic data},
  author={Fu, Stephanie and Tamir, Netanel and Sundaram, Shobhita and Chai, Lucy and Zhang, Richard and Dekel, Tali and Isola, Phillip},
  journal={arXiv preprint arXiv:2306.09344},
  year={2023}
}

@inproceedings{wang2026fair,
  title     = {Fair and Faithful: A Diffusion-Enhanced Dataset and Hybrid State-Space Mamba for Face Super-Resolution},
  author    = {Wang, Tao and Xia, Peiwen and Tang, Bowen and Chen, Jinwei and Zhang, Kaihao and Li, Bi},
  booktitle = {European Conference on Computer Vision},
  year      = 2026,
  pages     = {281--300},
}

@inproceedings{tang2026evolutionary,
  title     = {Evolutionary Multi-Agent Collaboration for Real-World Video Face Restoration},
  author    = {Tang, Bowen and Wang, Tao and Zhang, Miao and Yu, Xin and Chen, Jinwei and Li, Bo and Zhang, Kaihao},
  booktitle = {Proceedings of the IEEE/CVF Conference on Computer Vision and Pattern Recognition},
  year      = 2026,
  pages     = {8175--8184},
}

@article{pu2026drag,
  title={Drag as Evidence: Motion-Grounded Latent Recomposition for Drag-Based Editing},
  author={Pu, Xinyu and Wang, Hongsong and Gui, Jie and Zhou, Pan},
  journal={arXiv preprint arXiv:2609.36755},
  year={2026}
}

@inproceedings{pu2026dragging,
  title={Dragging with geometry: From pixels to geometry-guided image editing},
  author={Pu, Xinyu and Wang, Hongsong and Gui, Jie and Zhou, Pan},
  booktitle={International Conference on Learning Representations},
  volume={2026},
  pages={151958--151984},
  year={2026}
}

@inproceedings{yan2025mobileie,
  title={Mobileie: An extremely lightweight and effective convnet for real-time image enhancement on mobile devices},
  author={Yan, Hailong and Li, Ao and Zhang, Xiangtao and Liu, Zhe and Shi, Zenglin and Zhu, Ce and Zhang, Le},
  booktitle={2025 IEEE/CVF International Conference on Computer Vision (ICCV)},
  pages={21949--21960},
  year={2025},
  organization={IEEE}
}

@article{yan2025igdnet,
  title={Igdnet: Zero-shot robust underexposed image enhancement via illumination-guided and denoising},
  author={Yan, Hailong and Huang, Junjian and Huang, Tingwen},
  journal={IEEE Transactions on Artificial Intelligence},
  year={2025},
  publisher={IEEE}
}

@article{yan2026revisiting,
  title={Revisiting Lightweight Low-Light Image Enhancement: From a YUV Color Space Perspective},
  author={Yan, Hailong and Liu, Shice and Zhang, Xiangtao and Yao, Lujian and Yang, Fengxiang and Chen, Jinwei and Li, Bo},
  journal={arXiv preprint arXiv:2601.17349},
  year={2026}
}

@article{bai2026towards,
  title={Towards lightest low-light image enhancement architecture for mobile devices},
  author={Bai, Guangrui and Yan, Hailong and Liu, Wenhai and Deng, Yahui and Dong, Erbao},
  journal={Expert Systems with Applications},
  volume={296},
  pages={129125},
  year={2026},
  publisher={Elsevier}
}
}

\end{document}